\pdfoutput=1

\documentclass[11pt]{article}

\usepackage{subfig}
\usepackage{amsmath}
\usepackage{booktabs}
\usepackage{multirow}
\usepackage{graphicx}
\usepackage{amssymb}
\usepackage{amsmath}
\usepackage{amsthm}
\usepackage{colortbl}
\usepackage{algorithm}
\usepackage{algorithmic}

\usepackage[final]{acl}

\usepackage{times}
\usepackage{latexsym}

\usepackage[T1]{fontenc}

\usepackage[utf8]{inputenc}

\usepackage{microtype}

\usepackage{inconsolata}

\usepackage{graphicx}

\title{Mitigating Hallucination in Large Vision-Language Models through Aligning Attention Distribution to Information Flow}

\author{
 \textbf{Jianfei Zhao\textsuperscript{1,2}},
 \textbf{Feng Zhang\textsuperscript{1}},
 \textbf{Xin Sun\textsuperscript{1,}\footnotemark[1]},
 \textbf{Chong Feng\textsuperscript{1,3,}\thanks{Corresponding Authors.}} 
\\
 \textsuperscript{1}School of Computer Science and Technology, Beijing Institute of Technology\\
 \textsuperscript{2}Zhongguancun Academy\\
 \textsuperscript{3}Southeast Academy of Information Technology, Beijing Institute of Technology\\
 \small{
   \{zhqingan, bit\_zhangfeng, sunxin, fengchong\}@bit.edu.cn
 } 
}

\begin{document}

\maketitle
\begin{abstract}
Due to the unidirectional masking mechanism, Decoder-Only models propagate information from left to right. LVLMs (Large Vision-Language Models) follow the same architecture, with visual information gradually integrated into semantic representations during forward propagation. Through systematic analysis, we observe that the majority of the visual information is absorbed into the semantic representations. 
However, the model's attention distribution does not exhibit sufficient emphasis on semantic representations.
This misalignment between the attention distribution and the actual information flow undermines the model's visual understanding ability and contributes to hallucinations.
To address this issue, we enhance the model's visual understanding by leveraging the core information embedded in semantic representations. Specifically, we identify attention heads that focus on core semantic representations based on their attention distributions. Then, through a two-stage optimization paradigm, we propagate the advantages of these attention heads across the entire model, aligning the attention distribution with the actual information flow.
We evaluate our method on three image captioning benchmarks using five different LVLMs,
demonstrating its effectiveness in significantly reducing hallucinations. Further experiments reveal a trade-off between reduced hallucinations and richer details. Notably, our method allows for manual adjustment of the model's conservativeness, enabling flexible control to meet diverse real-world requirements.\footnote{Code is available at \url{https://github.com/beta-nlp/SEVI}.}
\end{abstract}

\section{Introduction}

\begin{figure}[tbp]
  \centering
  \includegraphics[width=1.0\linewidth]{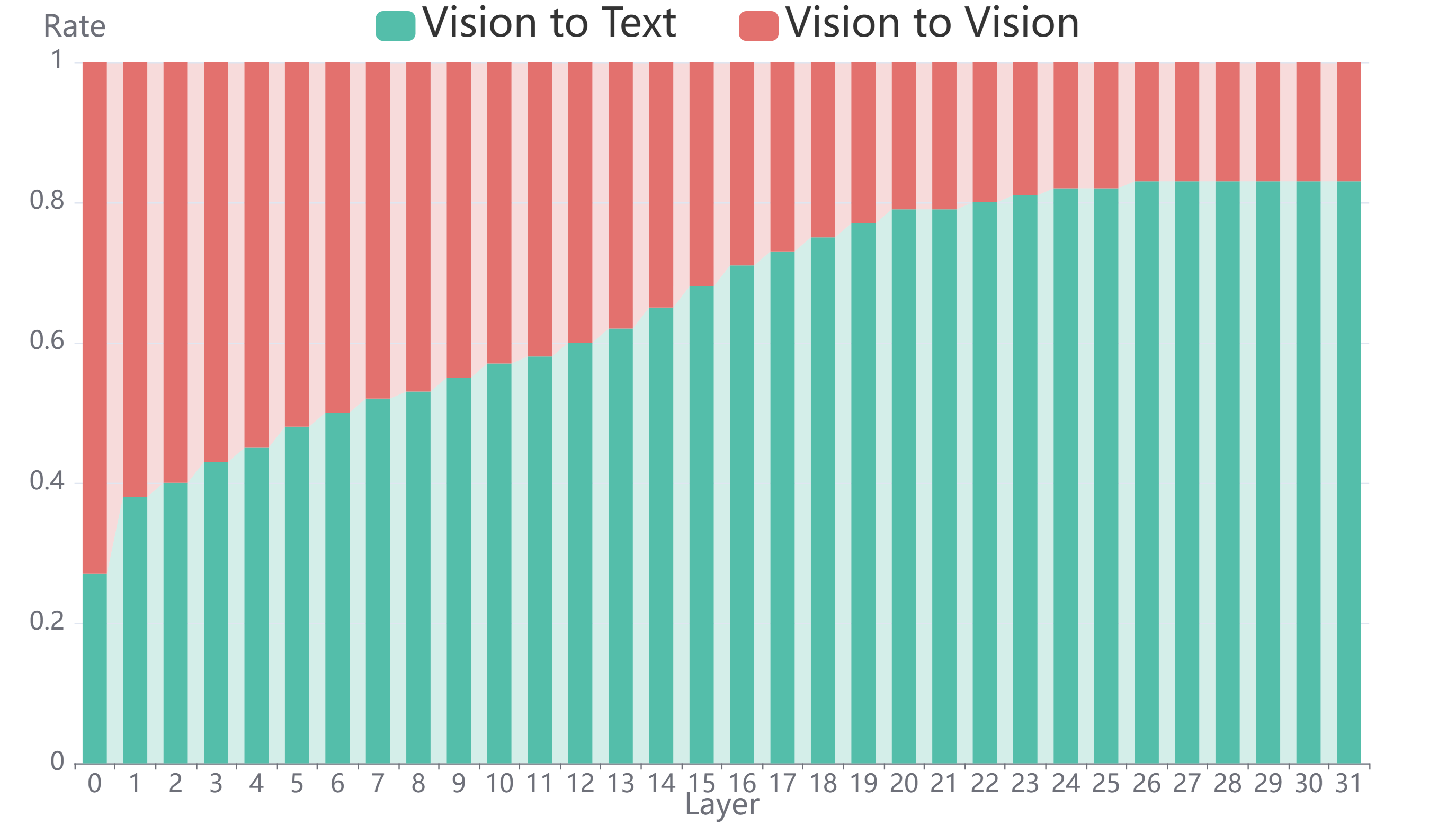}
  \caption{
    Visual information flow in the image captioning task on LLaVA-1.5.
  \emph{Vision to Vision} and \emph{Vision to Text} respectively denote
  the visual features' contributions to the visual representations and semantic representations.
  }
  \label{Fig_rollout}
\end{figure}
Large Vision-Language Models (LVLMs)~\cite{blip,qwen-vl,llava15} integrate Large Language Models (LLMs) with visual encoders, aligning the extracted visual features with the semantic space of the LLMs to enable comprehensive understanding of visual content. However, they often suffer from hallucination~\cite{wang2023evaluation,POPE,hallu_lvlm}, where the models may generate content that is inconsistent with the visual evidence, thereby severely limiting their reliability in real-world scenarios.

Recent studies have explored methods to mitigate hallucinations in LVLMs.~\citet{VCD} and~\citet{DeGF} attribute hallucinations to the influence of language priors, proposing contrastive decoding-based methods to suppress language priors. However, they do not explicitly enhance the model's visual understanding capabilities. Other studies~\cite{VAF,VAR,VHD} argue that LVLMs exhibit biased mechanisms in visual attention distribution, and propose optimizations such as increasing the relative weight of key visual tokens. In this work, however, we point out that these approaches make suboptimal adjustments to the information flow, as they overlook the influence of information aggregation.

In fact, the unidirectional masked generation process in Transformer-based models can be regarded as a form of information flow, where information from earlier tokens in the input or generated sequence flows toward later tokens. Since LVLMs adopt this architecture, their understanding of visual content can be modeled as a process in which visual information flows into the semantic representations (since visual tokens are positioned before the textual tokens). Based on this perspective, we follow the Attention Rollout~\cite{AttnFlow} to conduct an in-depth analysis of information flow within LVLMs. 
The visualization results are shown in Fig.~\ref{Fig_rollout}.

Specifically, we use $\mathbf{F}_{i,j}^{l}$ to represent the contributions of the $i\text{-th}$ input embedding to the $j\text{-th}$ representation in layer $l$.
Considering the residual connection in the forward propagation, we formulate the recursive relation as $\mathbf{F}^l=(\mathbf{I}+\mathbf{W}_\mathrm{attn})/2 \cdot \mathbf{F}^{l-1}$,
where $\mathbf{W}_\mathrm{attn}$ is the attention weight after average pooling across multi-heads.
The contributions of each input embedding were normalized to 1 across all representations.
We perform layer-wise quantification of visual embeddings' contributions to the visual and semantic representations, respectively.

As shown in the visualization, visual information progressively flows into semantic representations across layers. In the final layers, the majority of the visual information is integrated into the semantic representations. This indicates that \textbf{during the forward pass of the LVLMs, visual information is gradually encoded into the semantic representations}. While prior works largely concentrate on optimizing the attention distribution to visual features, they neglect the critical insight that essential information has already been integrated into semantic representations, where attention refinement may be more impactful.

\begin{figure}[tbp]
  \centering
  \includegraphics[width=1.0\linewidth]{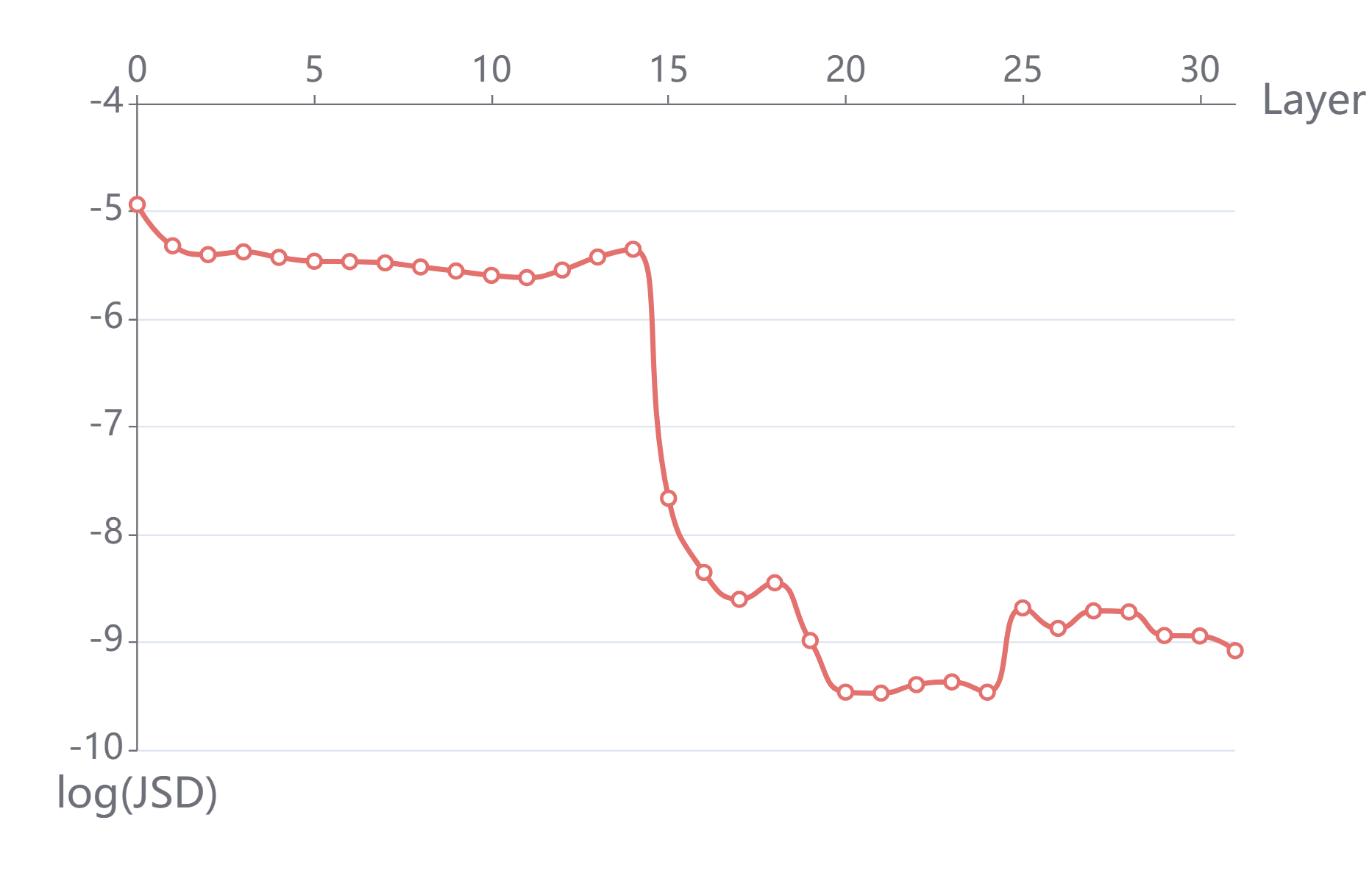}
  \caption{
    Jensen-Shannon divergence (JSD) between the regular and vision-masked logits distribution.
    \emph{layer=l} represents the application of masking vision features starting from the $l\text{-th}$ layer. 
    Lower $\log(\text{JSD})$ denotes higher consistency between two distributions.
    The generation context is \emph{The image depicts a \_}.
  }
  \label{Fig_mask_jsd}
\end{figure}

To further validate this insight, we mask all visual representations and compare the consistency between the logits distribution of the vision-masked input and that of the regular (unmasked) input, as illustrated in Fig.~\ref{Fig_mask_jsd}.
We observe that applying masking to visual representations starting from earlier layers leads to significant deviations in the model's outputs, whereas masking from later layers has minimal impact on the final outputs. This finding further supports our view that visual information has already been integrated into the semantic representations in the later layers.

\begin{figure}[tbp]
  \centering
  \includegraphics[width=1.0\linewidth]{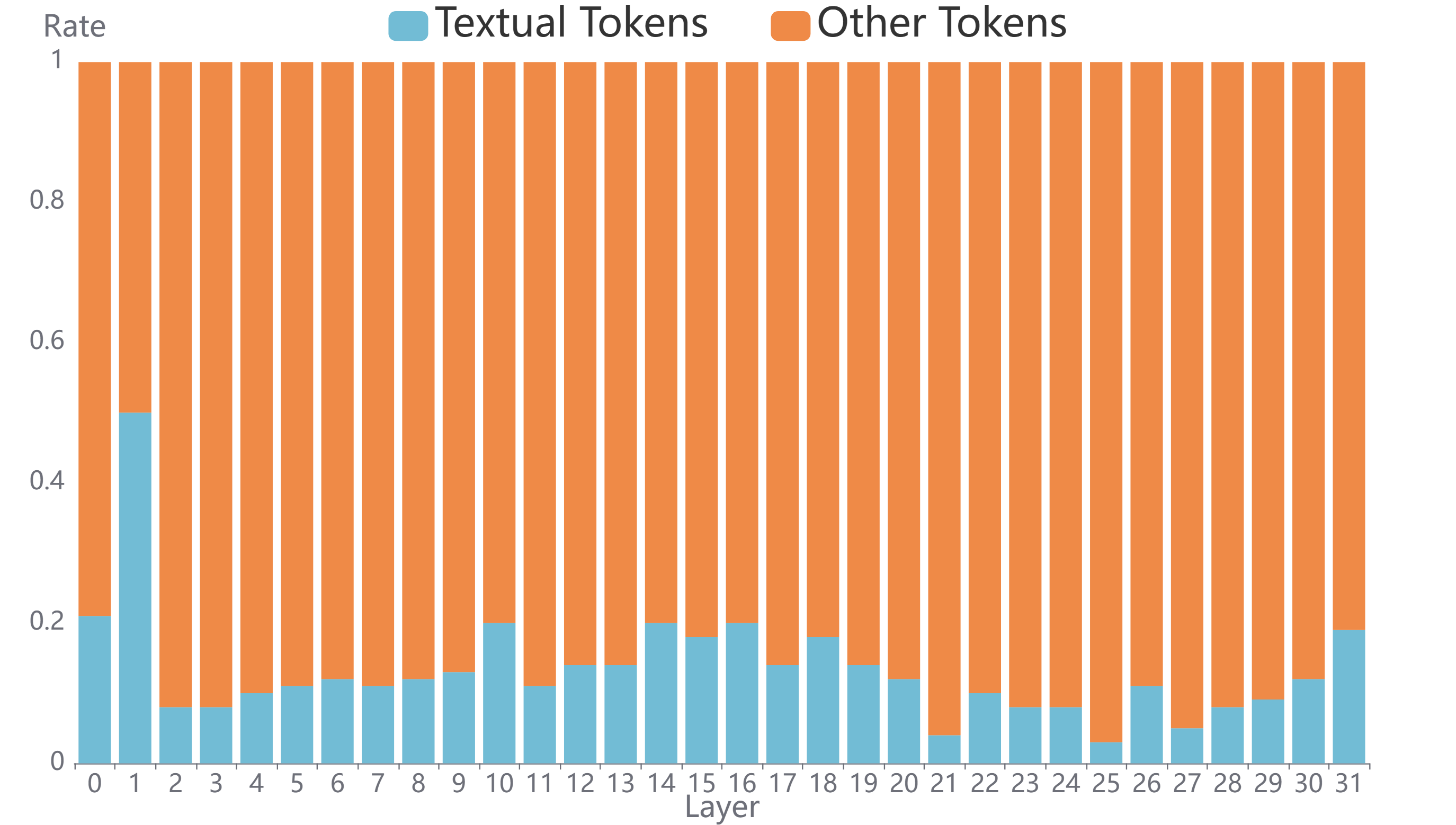}
  \caption{
    Attention distribution between semantic representations and all others, obtained by averaging across all attention heads.
  }
  \label{Fig_attn_vt}
\end{figure}

Building upon the above findings, as the number of layers increases, textual representations progressively integrate more visual information and thus play an increasingly important role. Consequently, LVLMs should place greater attention on these integrated representations.
To investigate this, we statistically analyze the attention allocation between semantic representations and others at each layer, as illustrated in Fig.~\ref{Fig_attn_vt}.
\textbf{The model's attention to semantic representations is significantly lower} than expected, accounting for only about one-fifth of the total attention. This phenomenon stems from the nature of supervision in LVLM training, which is primarily driven by soft supervision from end-to-end data and lacks explicit guidance on the attention distribution.
We argue that \textbf{the model's attention distribution should align with the actual flow pattern of visual information}. Otherwise, hallucinations may arise, as the model does not sufficiently attend to regions enriched with visual information.

To address the aforementioned challenge, we propose \textbf{SEVI} (Semantic-Enhanced Visual Interpretation), a novel training-free approach that augments the model's attention to those semantic representations that have absorbed visual information.
Furtherly, we encourage the model to focus on the most meaningful representations in order to reduce the occurrence of hallucinations.
Several studies~\cite{Anchors,StreamingLLM} have revealed that contextual information tends to spontaneously concentrate in some tokens, forming anchor tokens that encapsulate core information.
These tokens are typically characterized by receiving disproportionately high attention~\cite{Anchors}.
Therefore, our method focuses the model's attention on these core semantic representations at higher layers where information integration and aggregation occur.
The core visual information contained within these representations provides more effective guidance for the model's understanding of visual inputs, thereby mitigating hallucination.

Specifically, we identify superior attention heads within the multi-head attention mechanism—those that attend to core semantic representations—and use their attention distribution as the target distribution to guide the optimization of the model's overall attention.
To achieve this, we design a two-stage attention optimization paradigm, incorporating a smoothing mechanism to ensure stable performance improvements.

We evaluate the effectiveness of our method in mitigating hallucinations by testing it with five LVLMs: InstructBLIP~\cite{blip}, LLaVA-1.5~\cite{llava15}, LLaVA-Next~\cite{llava-next}, Qwen2-VL-Instruct~\cite{qwen2-vl}, and Qwen2.5-VL-Instruct~\cite{qwen25-vl}, 
on three image captioning benchmarks: CHAIR~\cite{chair}, AMBER~\cite{AMBER}, and DetailCaps~\cite{DetailCap}. 
Experimental results show that our approach significantly reduces hallucinations while supporting manual adjustment of the model's conservativeness, enabling a flexible trade-off between cautious outputs (fewer hallucinations) and comprehensive descriptions (richer details).


Our main contributions are as follows:
\begin{itemize}
  \item We reveal that visual information is indeed integrated into semantic representations. However, the model's attention distribution does not align with this flow pattern, leading to hallucinations.

  \item We enhance the visual understanding capabilities of LVLMs by leveraging semantic representations. Specifically, we guide the model's attention toward core semantic representations through a two-stage optimization paradigm, effectively reducing hallucinations.

  \item We evaluate our method on the image captioning task, demonstrating its effectiveness in significantly reducing hallucinations across three benchmark datasets and five LVLMs.
  We observe a trade-off between reducing hallucinations and preserving detail, with our method enabling controllable conservativeness for task-specific needs.
\end{itemize}

\section{Related Work}

Various approaches have been proposed to mitigate hallucinations in LVLMs, such as improving training data quality~\cite{LRV-Instruction,AugRLHF,DPHallu} or designing specific data formats~\cite{CIEM,HallE-Switch,PerturboLLaVA} to enhance model reliability.
However, these training-based methods often suffer from limited scalability.

In recent years, the rapid evolution of LVLM backbones has brought increasing attention to training-free methods.
Early approaches~\cite{Woodpecker,VIGC,AMOH} primarily relied on post-processing techniques to correct hallucinated content after generation.
However, the complexity of such pipelines and their dependence on external modules limit their practical usability.
Researchers have since shifted focus toward investigating the root causes of hallucinations, exploring targeted optimizations in both the decoding strategies and the decoding processes.
Some methods~\cite{VCD,ICD,DeGF,CICD} apply contrastive decoding algorithms to refine the logits, thereby reducing the likelihood of hallucination-related tokens. While these approaches improve the expressiveness of the model and enhance the transfer of visual information into textual form, they do not strengthen the model's intrinsic cross-modal understanding.
Other works have proposed enhancements to the attention mechanism, such as correcting visual positional bias~\cite{IMCCD,DAC}, directly boosting visual attention~\cite{IBD,VHD}, or redistributing attention weights~\cite{VAR,VAF}.
However, these methods primarily concentrate on visual attention and overlook the vision-integrated semantic representation.

In contrast to these approaches, we analyze the information flow between the visual and textual modalities in LVLMs and focus the model's attention on core information within semantic representations, thereby reducing hallucinations more effectively.

\section{Preliminaries}

LVLMs consist of a visual encoder and an LLM backbone.
High-dimensional features extracted by the visual encoder are concatenated with textual embeddings and jointly fed into the LLM.
Subsequently, the visual and semantic features undergo cross-modal interaction through the Self-Attention layers within the LLM.
The cross-modal interaction in layer $l$ simplified as $\Delta \mathbf{X}^l = \text{softmax}(\mathbf{W}) \mathbf{X}^l $,
where $\mathbf{X}^l = [\mathbf{X}^l_V: \mathbf{X}^l_T]$ denotes the hidden states of visual features $\mathbf{X}^l_V$ and semantic features $\mathbf{X}^l_T$.
$\mathbf{W} \in \mathbb{R}^{H \times L \times L} $ is the attention weights, $H$ is the number of attention head
and $L$ is the model's context length.

When generating the ($j$+1)-th token, the model's attention to feature $\mathbf{X}^l_i$ at layer $l$ across all heads is quantified by the weight value $\mathbf{W}_{i,j}$.
A larger $\mathbf{W}_{i,j}$ indicates that the model's current layer relies more heavily on the information from $\mathbf{X}^l_i$ when determining the generated content.
Therefore, we can intuitively infer that more important features $\mathbf{X}^l_i$ should generally correspond to larger attention weights $\mathbf{W}_{i,j}$.

\section{Method}
\label{sec:method}

\begin{figure*}
  \centering
  \includegraphics[width=\linewidth]{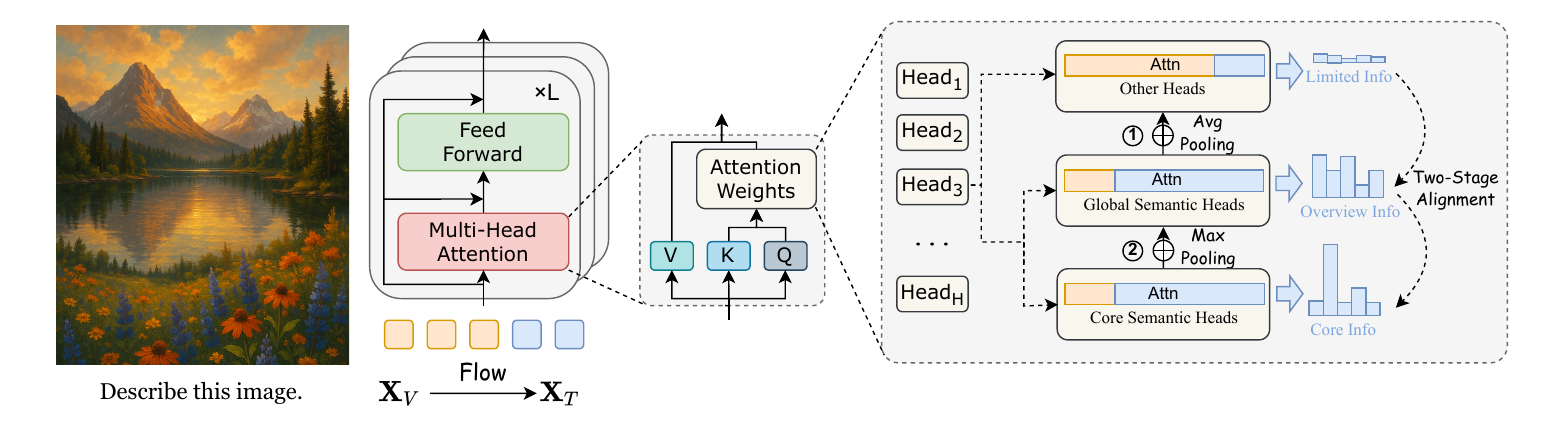}
  \caption{
    The diagram of the two-stage attention distribution alignment.
    We first categorize attention heads based on their focus on semantic representations into semantic heads and other heads.
    Semantic heads are further divided into core semantic heads and global semantic heads, depending on whether they attend to core semantic representations.
    We then align the model's attention distribution with the core semantic heads through a two-stage optimization process.
  }
  \label{Fig_diagram}
\end{figure*}

We attribute the occurrence of hallucinations to a misalignment between the attention distribution of LVLMs and the visual information flow.
To address this, we enhance the visual interpretation capabilities of LVLMs by leveraging semantic representations that encapsulate visual information.
We first identify superior attention heads that focus on core semantic representations, and then propagate their strengths to other heads through a two-stage optimization paradigm, as illustrated in Fig. \ref{Fig_diagram}.

\subsection{Attention Distribution Alignment}

Considering that visual information is progressively integrated into semantic representations through forward propagation—and that only a few of these representations aggregate core information—an intuitive approach is to increase the model's attention to these core semantic representations in the later layers.
However, rigidly modifying the model's attention distribution may disrupt its inherent latent features, potentially resulting in suboptimal performance.

In the multi-head attention mechanism, attention heads naturally diversify after training, forming a heterogeneous and specialized structure~\cite{transformer}.
Accordingly, we identify superior attention heads that attend to core semantic representations and transfer their attention patterns to other heads, leading to a global refinement of the model's attention distribution.

Specifically, at each layer, we calculate the attention weight of each head with respect to the semantic representations. We then identify those heads that allocate more than 50\% of their attention to semantic representations:
\begin{equation}
  h \in \left\{
\begin{aligned}
H_S \quad &\text{ if }  \sum \mathbf{W}^h_S > \sum \mathbf{W}^h_O \\
H_O \quad &\text{ if }  \sum \mathbf{W}^h_S \le \sum\mathbf{W}^h_O
\end{aligned}
  \right.
\end{equation}
$\mathbf{W}^h_S$ and $\mathbf{W}^h_S$ respectively denote the attention weights of head $h$ to semantic and other representations.
$H_S$ denotes the set of heads that focus more on semantic representations, whereas $H_O$ denotes other heads.
$H_S$ are further categorized based on their attention distributions into \emph{Core Semantic Heads} ($H_{S_c}$), which focus on core semantic representations, and \emph{Global Semantic Heads} ($H_{S_g}$), which fail to effectively attend to them:
\begin{equation} \label{Eq_kp}
  h \in \left\{
\begin{aligned}
H_{S_c} \quad &\text{if} \quad \max \mathbf{W}^h_S > \kappa \cdot \sum \mathbf{W}^h_S \\
H_{S_g} \quad &\text{if} \quad \max \mathbf{W}^h_S \le \kappa \cdot \sum \mathbf{W}^h_S
\end{aligned}
  \right.
\end{equation}
where $\kappa$ is a hyperparameter and $H_{S_c} \cup H_{S_g} = H_S$ holds.
We then take the core semantic heads to construct a target distribution to optimize the model's attention distribution.

To ensure stable information integration within the model,
we propose a two-stage feature optimization paradigm that more naturally guides the model's understanding of visual information.

In the first stage, we use the global semantic heads to guide the other heads, aligning the model's attention transition to semantic representations and harmonizing with the visual information flow.
In the second stage, we leverage the core semantic heads to guide the global semantic heads, encouraging the model to concentrate more on critical information and thereby suppressing hallucinations.

The model's internal attention is then optimized through a two-stage refinement process.

During this process, we incorporate a weighted smoothing mechanism to adjust the attention distribution of each head:

\begin{equation}
  \hat{\mathbf{W}}^h = \frac{\mathbf{W}^h + \omega \mathbf{W}'}{1+\omega}
\end{equation}
where $\hat{\mathbf{W}}^h$ denotes the optimized attention weights of head $h$, $\mathbf{W}'$ represents the target weight distribution derived from superior heads, and $\omega$ is a hyperparameter that controls the extent of the modification.

We first use the global semantic heads to guide the other heads,
aiming to align the model's attention distribution with the flow of visual information.
To this end, we apply average pooling across the global semantic heads to construct the target attention distribution:
\begin{equation}
   \hat{\mathbf{W}}^h_O = \frac{\mathbf{W}^h_O + \omega \cdot \frac{1}{|H_{S_g}|} \sum_{t \in H_{S_g}}  \mathbf{W}^t_{S_g}}{1+\omega}, 
\end{equation}
Subsequently, we use the core semantic heads to guide the global semantic heads, with the goal of enhancing the model's focus on critical information.
Here, we employ max pooling across core semantic heads to form the corresponding target distribution:
\begin{equation}
  \hat{\mathbf{W}}^h_{S_g} = \frac{\mathbf{W}^h_{S_g} + \omega \cdot \max_{t \in H_{S_c}} \mathbf{W}^t_{S_c}}{1+\omega}
\end{equation}

We provide the algorithm of the two-stage optimization process in Appendix~\ref{sec:alg}.

\begin{table*}[hbt!]
  \centering
  \begin{tabular}{lccccccccc}
    \toprule
    \multirow{2}{*}{\textbf{Method}} & \multicolumn{3}{c}{LLaVA-1.5}  & \multicolumn{3}{c}{InstructBLIP} & \multicolumn{3}{c}{LLaVA-1.5-13B}  \\
    \cmidrule(rl){2-4} \cmidrule(rl){5-7} \cmidrule(rl){8-10}
& \textbf{Cs}$\downarrow$ & \textbf{Ci}$\downarrow$ & \textbf{R} $\uparrow$  & \textbf{Cs}$\downarrow$ & \textbf{Ci}$\downarrow$ & \textbf{R}$\uparrow$  & \textbf{Cs}$\downarrow$ & \textbf{Ci}$\downarrow$ & \textbf{R} $\uparrow$   \\
    \midrule
    \multicolumn{10}{c}{\emph{Maximum Generation Length is 64}}  \\
    \midrule
    
Regular & 24.4 & 8.9 & 56.6 & 35.6 & 13.2 & 56.4 & 24.0 & 7.8 & 56.5     \\
VCD~\cite{VCD}  & 25.0 & 8.3 & 59.0 &  32.2 & 10.3 & 60.6 &  23.4 & 7.5 & 59.4                          \\
ICD~\cite{ICD} &  23.2 & 8.1 & 58.4 & 29.8 & 9.8 & 60.6  & 18.8 & 6.6 & 58.4                          \\
PAI~\cite{PAI} &  20.0 & 6.2 & 56.9 & \underline{26.0} & 8.9 & 53.9 &  \underline{17.4} & \underline{5.3} & 58.7     \\
IBD~\cite{IBD} &  21.2 & 6.9 & 58.8 & 27.8 & 9.2 & 60.3 & 22.4 & 7.1 & \textbf{59.7}                         \\
VAR~\cite{VAR} &  25.2 & 8.6 & 55.3 & - & - & - &  25.8 & 8.5 & 56.3\\
AD-HH~\cite{AD-HH} &  19.4 & 6.3 & 52.5 & - & - & - & 20.0 & 6.5 & 53.9\\
VAF~\cite{VAF} &  26.2 & 9.3 & 56.6 & 32.0 & 12.0 & 55.3  & 23.2 & 8.0 & 57.6\\
DeGF~\cite{DeGF} &  22.4 & 7.2 & 58.2 & 32.4 & 11.0 & 59.4 &  22.4 & 6.9 & \underline{59.6}                          \\
CICD~\cite{CICD} & \underline{18.0} & 6.1 & \textbf{59.6} & 23.8 & \underline{7.7} & \textbf{62.2} &  20.4 & 6.6 & 59.4\\
\rowcolor{gray!20}
SEVI$_{\mathrm{Balanced}}$ &  18.8 & \underline{5.5} & \underline{59.5} & \underline{27.8} & 8.8 & \underline{60.9} & 17.8 & 5.5 & 57.6                         \\
\rowcolor{gray!20}
SEVI$_{\mathrm{Focused}}$ &  \textbf{14.8} & \textbf{4.7} & 54.1 &  \textbf{15.4} & \textbf{5.9} & 50.9 & \textbf{13.4} & \textbf{4.5} & 54.4                         \\

    \midrule
    \multicolumn{10}{c}{\emph{Maximum Generation Length is 512}}  \\
    \midrule
Regular & 54.6 & 16.4 & 72.6 & 62.6 & 19.5 & 66.9 & 58.8 & 17.0 & 73.4    \\
VCD~\cite{VCD}  & 59.8 & 17.8 & \textbf{75.6}  & 64.8 & 18.8 & \textbf{71.9} & 60.2 & 16.4 & \textbf{76.9}                     \\
ICD~\cite{ICD} &  57.0 & 15.0 & 74.6  & 59.0 & 17.1 & 69.2 &  55.0 & 14.5 & \underline{76.4}                  \\
PAI~\cite{PAI} &  41.2 & 10.4 & 68.6 & 67.6 & 19.4 & 68.0 & 35.4 & 9.5 & 73.1                \\
IBD~\cite{IBD} &  57.6 & 16.5 & 74.2 & 57.6 & 15.7 & 70.8  & 50.6 & 14.3 & 76.1                  \\
VAR~\cite{VAR} &   60.0 & 18.3 & 72.6 & - & - & - &  54.8 & 15.3 & 73.8           \\
AD-HH~\cite{AD-HH} & 46.6 & 12.6 & 66.5 & - & - & - & 50.0 & 13.5 & 71.0                 \\
VAF~\cite{VAF} &  58.8 & 19.0 & 71.3 &  58.6 & 17.8 & 66.9 & 57.2 & 16.2 & 73.7\\
DeGF~\cite{DeGF} &  57.4 & 16.3 & \underline{75.5}  & 59.0 & 17.7 & \underline{71.5} & 51.8 & 14.2 & 75.5                    \\
CICD~\cite{CICD} & 43.8 & 11.7 & 75.0 & 49.8 & 13.7 & 70.3 & 44.4 & 12.9 & 76.0                  \\
    \rowcolor{gray!20}
SEVI$_{\mathrm{Balanced}}$ &  \underline{34.8} & \underline{9.0} & 68.3 & \underline{41.0} & 12.4 & 67.1 & \underline{28.6} & \underline{8.4} & 68.7                         \\
    \rowcolor{gray!20}
SEVI$_{\mathrm{Focused}}$ &  \textbf{17.8} & \textbf{5.5} & 56.9 & \textbf{18.8} & \textbf{8.4} & 51.6 & \textbf{15.0} & \textbf{5.1} & 55.6                           \\

    \bottomrule
  \end{tabular}
  \caption{
      Results on  CHAIR.
      SEVI$_{\mathrm{Balanced}}$ and SEVI$_{\mathrm{Focused}}$ represent our method operating in the \emph{Balanced} and \emph{Focused} mode, respectively.
      \textbf{Cs} and \textbf{Ci} represent CHAIRs and CHAIRi, \textbf{R} represents Recall.
      }\label{Tb_chair}
\end{table*}
\subsection{Mitigating Aggravated Language Priors}
We optimize the model's attention distribution to align with the flow pattern of visual information,
which ultimately results in increased attention to semantic representations.
However, while semantic representations incorporate visual information, they also inherently contain their own contextual features.
As a result, increasing the model's reliance on semantic representations will exacerbate the issue of language priors~\cite{wu2022overcoming,ren2023overcoming,PerturboLLaVA}.

To address this issue, we employ CICD~\cite{CICD} to eliminate language priors while preserving visual information.
CICD uses the cross-image consistency of language priors to identify detrimental priors and mitigate them by contrastive decoding:
\begin{equation}
  \begin{aligned}
      \mathrm{logit}(y_t \mid v, x, y_{<t}) =\;& (1 + \alpha)\, \hat{\mathrm{logit}}_{\theta}(y_t \mid v, x, y_{<t}) \\
  & - \alpha\, \mathrm{logit}_{\theta}(y_t \mid v', x, y_{<t})
  \end{aligned}
\end{equation}
where $\hat{\mathrm{logit}}_{\theta}$ is the logit distribution after attention distribution alignment
and $\mathrm{logit}_{\theta}$ is the regular logit distribution with a distinct image $v'$.
$\alpha$ is defined as follows:
\begin{equation}
\alpha = 1 - \log_{10}(\mathrm{JSD}(\hat{\mathrm{logit}}_{\theta}; \mathrm{logit}_{\theta}))
\end{equation}
where $\mathrm{JSD}(\cdot; \cdot)$ represents Jensen-Shannon Divergence.

\section{Experiments}

\begin{table*}[t]
  \centering
  \begin{tabular}{lcccccc}
    \toprule
    \multirow{2}{*}{\textbf{Method}} & \multicolumn{3}{c}{LLaVA-1.5} & \multicolumn{3}{c}{InstructBLIP} \\
    \cmidrule(rl){2-4} \cmidrule(rl){5-7}
    
    & \textbf{CAPTURE}$\uparrow$ & \textbf{Cs}$\downarrow$ & \textbf{Ci}$\downarrow$  & \textbf{CAPTURE}$\uparrow$ & \textbf{Cs}$\downarrow$ & \textbf{Ci}$\downarrow$ \\
     
     \midrule
      Regular &  52.60 &  55.7 & 17.4  & 52.99 & 58.6 & 18.0   \\
      VCD~\cite{VCD}  &  52.91  &  55.7 & 16.8   & 53.20 &  59.6 & 18.9    \\
      ICD~\cite{ICD} &  52.82  &  53.9 & 16.6   & 53.24 &  55.7 & 16.7      \\
      PAI~\cite{PAI} &  53.49 & 39.4 & 11.8 & 53.27 & 62.6 & 18.3     \\
      IBD~\cite{IBD} &  52.48 &  54.1 & 15.8  & 54.14 & 56.4 & 15.5     \\
      VAR~\cite{VAR} & 52.98  & 55.1 & 17.1 & -  & - & -   \\
      AD-HH~\cite{AD-HH} &  52.52 & 44.4 & 11.5 & -  & - & -   \\
      VAF~\cite{VAF} & 52.36  & 55.6 & 18.1 & 52.63  & 55.9 & 16.9 \\
      DeGF~\cite{DeGF} & 52.72 &  55.7 & 16.6   & 53.06 &  59.1 & 17.4  \\
      CICD~\cite{CICD} & \underline{55.80}  &  45.6 & 13.1   & \underline{54.20} & 45.7 & 13.2   \\
    \rowcolor{gray!20}
      SEVI$_{\mathrm{Balanced}}$ & 53.73  & \underline{31.9} & \underline{8.8}  & 53.33  & \underline{38.0} & \underline{12.8}   \\
    \rowcolor{gray!20}  
      SEVI$_{\mathrm{Focused}}$ & \textbf{58.00}  & \textbf{19.3} & \textbf{6.1}   & \textbf{56.89}  & \textbf{18.3} & \textbf{8.2}  \\
    \bottomrule
  \end{tabular}
  \caption{
      Results on the COCO subset of DetailCaps.
      The maximum generation length is 512.
      }\label{Tb_DC}
\end{table*}

\begin{table*}[hbt!]
  \centering
  \setlength{\tabcolsep}{1mm}
  \begin{tabular}{lcccccccc}
    \toprule
     \multirow{2}{*}{\textbf{Method}} & \multicolumn{4}{c}{LLaVA-1.5} & \multicolumn{4}{c}{InstructBLIP}  \\
    \cmidrule(rl){2-5}\cmidrule(rl){6-9}
     & \textbf{CHAIR}$\downarrow$ & \textbf{Cover}$\uparrow$ & \textbf{Hal}$\downarrow$ & \textbf{Cog} $\downarrow$ & \textbf{CHAIR}$\downarrow$ & \textbf{Cover}$\uparrow$ & \textbf{Hal}$\downarrow$ & \textbf{Cog} $\downarrow$  \\
     \midrule
      Regular & 11.6 & 49.7 & 47.7 & 4.4 & 12.4 & 51.9 & 52.4 & 5.0   \\
      VCD~\cite{VCD}  &  9.8 & 51.2 & 43.8 & 4.4  &    9.9 & 54.0 & 44.6 & 4.2                  \\
      ICD~\cite{ICD}  &  8.8 &\underline{ 51.2} & 38.7 & 4.1  & 9.8 & 53.9 & 46.7 & 5.1                \\
      PAI~\cite{PAI}  &  7.7 & 49.3 & 36.9 & 3.3 & 11.7 & 52.8 & 55.1 & 5.4                  \\
      IBD~\cite{IBD}  &  9.8 & 50.5 & 42.2 & 4.4 & 9.0 & \textbf{56.1} & 45.1 & 4.6                  \\
      PAI~\cite{PAI}  &  7.7 & 49.3 & 36.9 & 3.3 & 11.7 & 52.8 & 55.1 & 5.4                  \\
      VAR~\cite{VAR} &  11.7 & \underline{51.2} & 48.5 & 4.8  & - & - & - & -                      \\
      AD-HH~\cite{AD-HH} &  9.0 & 48.0 & 40.8 & 3.0   & - & - & - & -                      \\
       VAF~\cite{VAF} & 11.3 & 50.2 & 48.6 & 4.3 & 11.5 & 51.8 & 50.1 & 5.1                \\
       DeGF~\cite{DeGF} &  9.1 & 50.7 & 39.9 & 4.1   &  9.7 & \underline{54.1} & 44.5 & 5.2                   \\
       CICD~\cite{CICD} &  6.6 & \textbf{52.7} & 34.8 & 2.2  & \underline{7.1} & 53.6 & 35.0 & 2.3                   \\
    \rowcolor{gray!20}
      SEVI$_{\mathrm{Balanced}}$ &  \textbf{5.6} & 48.6 & \underline{27.6} & \underline{1.7} & \textbf{6.0} & 51.0 & \underline{28.5} & \underline{1.6}                    \\
    \rowcolor{gray!20}
       
       SEVI$_{\mathrm{Focused}}$ & \underline{6.1} & 42.3 & \textbf{20.2} & \textbf{0.8} & 7.7 & 42.8 & \textbf{24.9} & \textbf{1.2}                    \\
    \bottomrule
  \end{tabular}
  \caption{
      Results on AMBER. The maximum generation length is 512.
    }\label{Tb_amber}
\end{table*}

\subsection{Experimental Setup}
Due to space constraints, we present only the key
aspects of our experimental setup here. Detailed
settings can be found in Appendix \ref{Sec:Detailed Experiments}.

\paragraph{Implementation Details}
Our method guides the model to focus on core information, thereby reducing the occurrence of hallucinations. In practice, we observe a trade-off between hallucination suppression and the richness of generated details, influenced by the model's degree of attentional focus. To address this, we introduce hyperparameters to control the model's focus on the core information.
Based on empirical observations, we design two sets of hyperparameter configurations: (1) \emph{Focused} mode that aggressively minimizes hallucinations, and (2) \emph{Balanced} mode that strikes a compromise between detail retention and hallucination reduction.
Specifically, \emph{Focused} mode optimizes attention distribution starting from the 5th layer with $\omega = 4$, while \emph{Balanced} mode begins from the 9th layer with $\omega = 0.5$.

In addition, we set the parameter $\kappa$ in Eq. \ref{Eq_kp} to 0.2. Further details are discussed in Appendix \ref{Sec_hy}.
We employ sampling decoding for the next-token prediction with default settings.
All experiments are performed on a single NVIDIA A800 40G GPU.

\paragraph{Benchmarks}
We evaluate the performance of our method across three widely adopted multimodal hallucination benchmarks on the image captioning task.
These include CHAIR~\cite{chair}, DetailCaps~\cite{DetailCap}, and AMBER~\cite{AMBER}.
\paragraph{Evaluated LVLMs}
To examine the generalizability of our approach, we apply it to five LVLMs drawn from three representative model families: InstructBLIP~\cite{blip}; two models from the LLaVA family (LLaVA-1.5~\cite{llava15} and LLaVA-Next~\cite{llava-next}); and two from the Qwen series (Qwen2-VL-Instruct~\cite{qwen2-vl} and Qwen2.5-VL-Instruct~\cite{qwen25-vl}). All models are tested at the 7B parameter scale unless explicitly noted otherwise.
\paragraph{Baselines}
We conduct a comparison between our method and several SOTA de-hallucination techniques: VCD~\cite{VCD}, ICD~\cite{ICD}, IBD~\cite{IBD}, PAI~\cite{PAI}, VAR~\cite{VAR}, VAF~\cite{VAF}, AD-HH~\cite{AD-HH}, DeGF~\cite{DeGF}, and CICD~\cite{CICD}. 

\begin{table*}[hbt!]
  \centering
  \begin{tabular}{lccccccccc}
    \toprule
    \multirow{2}{*}{\textbf{Method}} & \multicolumn{3}{c}{LLaVA-Next} & \multicolumn{3}{c}{Qwen2-VL} & \multicolumn{3}{c}{Qwen2.5-VL} \\
    \cmidrule(rl){2-4} \cmidrule(rl){5-7}\cmidrule(rl){8-10}
    & \textbf{Cs}$\downarrow$ & \textbf{Ci}$\downarrow$ & \textbf{R}$\uparrow$ & \textbf{Cs}$\downarrow$ & \textbf{Ci}$\downarrow$ & \textbf{R}$\uparrow$ & \textbf{Cs}$\downarrow$ & \textbf{Ci}$\downarrow$ & \textbf{R}$\uparrow$    \\
     \midrule
      Regular &   32.2 & 11.0 & \textbf{56.4} & 30.2 & 8.3 & 53.1 & 28.6 & 9.3 & \textbf{56.0}      \\
      \cmidrule(rl){1-10}
        SEVI$_{\mathrm{Balanced}}$ &   25.8 & 9.3 & 49  & 20.2 & \textbf{5.3} & \textbf{54.3} & 22.2 & \textbf{7.0} & 53.9                         \\
       SEVI$_{\mathrm{Focused}}$ & \textbf{20.8} & \textbf{9.9} & 37.8  & \textbf{18.8} & 7.0 & 41.9 &  \textbf{16.4} & 7.1 & 39.2                            \\
    \bottomrule
  \end{tabular}
  \caption{
      Results CHAIR with more LVLMs.
      The maximum generation length is 128.
    }\label{Tb_more}
\end{table*}

\subsection{Main Results}
\paragraph{Results on CHAIR}
CHAIR is a benchmark designed to detect object hallucinations in image captions, relying on human annotations to provide reliable ground truth.
Following common practice, we conduct experiments with maximum sequence lengths set to 64 and 512, respectively.
The results are shown in Fig. \ref{Tb_chair}.
Our method effectively reduces hallucinations by guiding the model's attention toward core information, achieving particularly low hallucination rates under the focused mode. However, we observe that when the model concentrates on critical information, it tends to become more conservative, as reflected by a slight drop in recall. In contrast, the balanced mode allows the model to capture more details while still maintaining a low hallucination rate, resulting in the best overall performance.
In addition, our method yields the most pronounced improvement on the 13B model relative to the 7B model, indicating its effectiveness in harnessing the latent capacity of larger models for visual understanding.

\begin{table}[t]
  \centering

  \begin{tabular}{lccc}
    \toprule
    \textbf{Setting}   & \textbf{Cs} $\downarrow$ & \textbf{Ci} $\downarrow$   & \textbf{R} $\uparrow$   \\
    \midrule
    \emph{PAI with different layers} &    \\
    None (only CD) &  25.4 & 7.6 & 61.9     \\
    $[${2-15}$]$ (Lower) &  20.8 & 6.8 & 56.7       \\
    $[${16-31}$]$ (Higher) & 24.6 & 8.0 & 59.3       \\
    $[${2-31}$]$ (PAI) & \textbf{20.0} & \textbf{6.0} & 57.6    \\
    \midrule
    \emph{SEVI combined with PAI} &    \\
    Regular & 24.4 & 8.9 & 56.6     \\
      \cmidrule(rl){1-4}
    SEVI$_{\mathrm{Focused}}$ & 14.8  & \textbf{4.7} & 54.1       \\
    SEVI$_{\mathrm{Focused}}$ \emph{w/} PAI & \textbf{14.2} & 5.3 & 52.2      \\
      \cmidrule(rl){1-4}
    SEVI$_{\mathrm{Balanced}}$ & 18.8 & 5.5 & 59.5    \\
    SEVI$_{\mathrm{Balanced}}$  \emph{w/} PAI & \textbf{15.0} & \textbf{5.3} & 55.4 \\
    \bottomrule
  \end{tabular}
  \caption{
      PAI increase visual attention from the 2nd layer to the final layer ($[${2-31}$]$).
      \emph{None (only CD)} denotes the use of contrastive decoding alone, without increasing visual attention.
      }\label{Tb_PAI}
\end{table}

\begin{table}[t]
  \centering
  \setlength{\tabcolsep}{1mm}
  
  \begin{tabular}{lcccc}
    \toprule
    \multirow{2}{*}{\textbf{Setting}} & \multicolumn{2}{c}{LLaVA-1.5} & \multicolumn{2}{c}{InstructBLIP} \\
      \cmidrule(rl){2-3}\cmidrule(rl){4-5}
    & \textbf{Cs}$\downarrow$ & \textbf{Ci}$\downarrow$   & \textbf{Cs}$\downarrow$ & \textbf{Ci}$\downarrow$  \\
    \midrule
    Regular &  54.6 & 16.4  & 62.6 & 19.5      \\
    \emph{w/o} CICD &  38.6 & 13.9 & 44.4 & 20.3       \\
    \emph{w/o} Core Heads & 39.0 & 9.7 & 45.0 & 13.3       \\
    \emph{w/o} Global Heads & 22.2 & 7.0 & 23.2 & 11.6       \\
    \emph{w/o} Two-Stage Opt. & 18.4 & 9.6 & 35.2 & 10.7  \\ 
      \cmidrule(rl){1-5}
    SEVI$_{\mathrm{Focused}}$  & \textbf{17.8} & \textbf{5.5}  & \textbf{18.8} & \textbf{8.4}       \\
    \bottomrule
  \end{tabular}
  \caption{\label{Tb_abla}
      Ablation study on CHAIR with \emph{Focused} mode.
      \emph{w/o Two-Stage Opt.} represents applying core semantic heads to non-semantic heads. We find that this approach disrupts the model's internal representations and leads to response collapse. To improve the smoothness of attention alignment, we set $\omega=1$ in this setting.
      }
\end{table}

\paragraph{Results on DetailCaps}
DetailCaps evaluates the correctness of captions in terms of objects, attributes, and relationships, incorporating both exact-match and soft-match metrics. The experimental results are presented in Tab. \ref{Tb_DC}.
Our method outperforms existing approaches in both hallucination rate and overall caption correctness. Moreover, the focused mode achieves a higher CPAURE score, indicating not only a lower incidence of hallucinations but also greater accuracy in the described content. This highlights the practical value of our approach for real-world applications.

\paragraph{Results on AMBER}
AMBER carefully selects high-quality images to construct its benchmark and provides a more fine-grained evaluation of object hallucinations. The experimental results are shown in Tab. \ref{Tb_amber}.
Our method achieves a significantly lower hallucination rate while maintaining a comparable Cover score to other approaches, resulting in the best overall performance.

\subsection{Visual or Semantic Representations, Which Should LVLMs Focus On?}
Some works~\cite{PAI,IBD} emphasize the importance of visual attention, whereas we encourage the model to focus more on semantic representations. To investigate whether the model should prioritize visual or semantic representations, we take PAI~\cite{PAI} as a contrast, which directly increases the model's attention to visual representations.

As we analyze in Fig.~\ref{Fig_rollout} and Fig.~\ref{Fig_mask_jsd}, visual information resides in visual representations during the early layers of LVLMs and is gradually integrated into semantic representations in later layers.
Therefore, we hypothesize that \textbf{LVLMs should focus on visual representations in the lower layers and shift to semantic representations in the higher layers}.
We perform ablation studies on both lower and higher layers within the PAI to better understand the layer-wise impact.
The results at the top of Tab.~\ref{Tb_PAI} are consistent with our hypothesis, indicating that it is more beneficial for the model to focus on visual representations in the lower layers.
Furthermore, we explore combining PAI with our method by applying PAI to the lower layers (layers [2–15]) to enhance visual attention, while adopting our approach in the higher layers (layers [16–31]).
The results at the bottom of Tab.~\ref{Tb_PAI} further demonstrate our hypothesis.

\subsection{Effectiveness on more LVLMs}
We evaluated our method on several state-of-the-art LVLMs, with the experimental results presented in Tab. \ref{Tb_more}.
Across these models, our method continues to demonstrate strong hallucination suppression capabilities, with both modes exhibiting their expected effectiveness. These results validate the excellent generalization ability of our approach and highlight its practical potential as a training-free solution.

\subsection{Ablation Study}
Through a two-stage optimization paradigm, we reallocate the model's attention toward the semantic representations, encouraging it to focus on the core visual information and thereby reducing hallucinations. In addition, we employ CICD to address the issue of language priors that aggravate when the model over-focuses on semantic representations.
We conduct ablation studies on these three components, with the results presented in Tab. \ref{Tb_abla}.
The core semantic heads play a pivotal role in guiding the model's focus, and removing them in the ablation study leads to a noticeable increase in hallucination rate. The global semantic heads help modulate the cross-modal attention distribution, aligning it with the flow of visual information, which also contributes significantly to hallucination reduction.
The combination of both components has a synergistic effect, and incorporating the CICD method to mitigate language priors, amplified by overattending to semantic representations, further enhances the performance of the model.
Additionally, the two-stage optimization paradigm also has a positive effect, making the optimization process more stable and smooth.
The experimental results confirm the effectiveness and soundness of our method's design, highlighting the value of each component.

\section{Conclusion}
We systematically analyze the information flow in LVLMs and verify that visual information is indeed integrated into the semantic representations. However, the model's attention remains predominantly focused on the visual representation. This inconsistency impairs the model's visual understanding and contributes to hallucination. To address this, we propose Semantic-Enhanced Visual Interpretation (SEVI), a method that guides the model's attention toward the core components of semantic representations through a two-stage optimization process. Extensive experiments demonstrate that our approach significantly mitigates hallucinations.

While our method optimizes the attention distribution in LVLMs based on the underlying information flow, it does not directly enhance the flow mechanism itself. In future work, we plan to explore more efficient strategies for visual information propagation.

\section*{Limitations}
Our method guides the model to focus on the most critical information, thereby reducing the occurrence of hallucinations. During our experiments, we observed a trade-off mechanism in the model's focusing process: excessive focus leads to a more conservative image captioning. In an effort to avoid hallucinated content, the model may overlook some details, which is reflected in a slight decrease in recall. 
Although our method demonstrates superior overall performance—showing a significant advantage on comprehensive metrics such as CAPTURE—it still faces a trade-off between reducing hallucinations and generating more details. To address this issue, we introduce hyperparameters to control the model's level of conservativeness, allowing users to manually adjust the behavior based on specific application scenarios. Furthermore, as a training-free approach, our method offers greater usability but is inherently limited by the performance ceiling of the model itself.

\section*{Acknowledgments}
We thank all reviewers for their detailed reviews and valuable comments.
This work is supported by Zhongguancun Academy Project No.20240103.
\bibliography{custom}

\begin{thebibliography}{43}
\providecommand{\natexlab}[1]{#1}

\bibitem[{Abnar and Zuidema(2020)}]{AttnFlow}
Samira Abnar and Willem Zuidema. 2020.
\newblock Quantifying attention flow in transformers.
\newblock In \emph{Proc. of ACL}, pages 4190--4197.

\bibitem[{Bai et~al.(2023)Bai, Bai, Yang, Wang, Tan, Wang, Lin, Zhou, and Zhou}]{qwen-vl}
Jinze Bai, Shuai Bai, Shusheng Yang, Shijie Wang, Sinan Tan, Peng Wang, Junyang Lin, Chang Zhou, and Jingren Zhou. 2023.
\newblock Qwen-vl: A versatile vision-language model for understanding, localization, text reading, and beyond.
\newblock \emph{arXiv preprint arXiv:2308.12966}.

\bibitem[{Bai et~al.(2025)Bai, Chen, Liu, Wang, Ge, Song, Dang, Wang, Wang, Tang et~al.}]{qwen25-vl}
Shuai Bai, Keqin Chen, Xuejing Liu, Jialin Wang, Wenbin Ge, Sibo Song, Kai Dang, Peng Wang, Shijie Wang, Jun Tang, and 1 others. 2025.
\newblock {Qwen2.5-vl} technical report.
\newblock \emph{arXiv preprint arXiv:2502.13923}.

\bibitem[{Chen et~al.(2025)Chen, Liu, Jing, Zhou, Rao, Chen, Zhang, and Shen}]{PerturboLLaVA}
Cong Chen, Mingyu Liu, Chenchen Jing, Yizhou Zhou, Fengyun Rao, Hao Chen, Bo~Zhang, and Chunhua Shen. 2025.
\newblock Perturbollava: Reducing multimodal hallucinations with perturbative visual training.
\newblock In \emph{Proc. of ICLR}.

\bibitem[{Dai et~al.(2023)Dai, Li, Li, Tiong, Zhao, Wang, Li, Fung, and Hoi}]{blip}
Wenliang Dai, Junnan Li, Dongxu Li, Anthony Tiong, Junqi Zhao, Weisheng Wang, Boyang Li, Pascale~N Fung, and Steven Hoi. 2023.
\newblock Instructblip: Towards general-purpose vision-language models with instruction tuning.
\newblock In \emph{Proc. of NIPS}, pages 49250--49267.

\bibitem[{Gunjal et~al.(2024)Gunjal, Yin, and Bas}]{DPHallu}
Anisha Gunjal, Jihan Yin, and Erhan Bas. 2024.
\newblock Detecting and preventing hallucinations in large vision language models.
\newblock In \emph{Proc. of AAAI}, pages 18135--18143.

\bibitem[{He et~al.(2025)He, Zhu, Guo, Fang, Hua, Jia, Tang, Chua, and Wang}]{VHD}
Jinghan He, Kuan Zhu, Haiyun Guo, Junfeng Fang, Zhenglin Hua, Yuheng Jia, Ming Tang, Tat-Seng Chua, and Jinqiao Wang. 2025.
\newblock Cracking the code of hallucination in lvlms with vision-aware head divergence.
\newblock In \emph{Proc. of ACL}.

\bibitem[{Hu et~al.(2023)Hu, Zhang, Zhao, and Sun}]{CIEM}
Hongyu Hu, Jiyuan Zhang, Minyi Zhao, and Zhenbang Sun. 2023.
\newblock Ciem: Contrastive instruction evaluation method for better instruction tuning.
\newblock In \emph{NIPS Workshop on Instruction Tuning and Instruction Following}.

\bibitem[{Kang et~al.(2025)Kang, Kim, Kim, and Hwang}]{VAR}
Seil Kang, Jinyeong Kim, Junhyeok Kim, and Seong~Jae Hwang. 2025.
\newblock See what you are told: Visual attention sink in large multimodal models.
\newblock In \emph{Proc. of ICLR}.

\bibitem[{Leng et~al.(2024)Leng, Zhang, Chen, Li, Lu, Miao, and Bing}]{VCD}
Sicong Leng, Hang Zhang, Guanzheng Chen, Xin Li, Shijian Lu, Chunyan Miao, and Lidong Bing. 2024.
\newblock Mitigating object hallucinations in large vision-language models through visual contrastive decoding.
\newblock In \emph{Proc. of CVPR}, pages 13872--13882.

\bibitem[{Li et~al.(2025)Li, Zhang, Jie, Ma, and Li}]{IMCCD}
Jiaming Li, Jiacheng Zhang, Zequn Jie, Lin Ma, and Guanbin Li. 2025.
\newblock Mitigating hallucination for large vision language model by inter-modality correlation calibration decoding.
\newblock \emph{arXiv preprint arXiv:2501.01926}.

\bibitem[{Li et~al.(2023)Li, Du, Zhou, Wang, Zhao, and Wen}]{POPE}
Yifan Li, Yifan Du, Kun Zhou, Jinpeng Wang, Wayne~Xin Zhao, and Ji-Rong Wen. 2023.
\newblock Evaluating object hallucination in large vision-language models.
\newblock In \emph{Proc. of EMNLP}, pages 292--305.

\bibitem[{Lin et~al.(2014)Lin, Maire, Belongie, Hays, Perona, Ramanan, Doll{\'a}r, and Zitnick}]{MSCOCO}
Tsung-Yi Lin, Michael Maire, Serge Belongie, James Hays, Pietro Perona, Deva Ramanan, Piotr Doll{\'a}r, and C~Lawrence Zitnick. 2014.
\newblock Microsoft coco: Common objects in context.
\newblock In \emph{Proc. of ECCV}, pages 740--755.

\bibitem[{Liu et~al.(2024{\natexlab{a}})Liu, Lin, Li, Wang, Yacoob, and Wang}]{LRV-Instruction}
Fuxiao Liu, Kevin Lin, Linjie Li, Jianfeng Wang, Yaser Yacoob, and Lijuan Wang. 2024{\natexlab{a}}.
\newblock Mitigating hallucination in large multi-modal models via robust instruction tuning.
\newblock In \emph{Proc. of ICLR}.

\bibitem[{Liu et~al.(2024{\natexlab{b}})Liu, Xue, Chen, Chen, Zhao, Wang, Hou, Li, and Peng}]{hallu_lvlm}
Hanchao Liu, Wenyuan Xue, Yifei Chen, Dapeng Chen, Xiutian Zhao, Ke~Wang, Liping Hou, Rongjun Li, and Wei Peng. 2024{\natexlab{b}}.
\newblock A survey on hallucination in large vision-language models.
\newblock \emph{arXiv preprint arXiv:2402.00253}.

\bibitem[{Liu et~al.(2024{\natexlab{c}})Liu, Li, Li, and Lee}]{llava15}
Haotian Liu, Chunyuan Li, Yuheng Li, and Yong~Jae Lee. 2024{\natexlab{c}}.
\newblock Improved baselines with visual instruction tuning.
\newblock In \emph{Proc. of CVPR}, pages 26296--26306.

\bibitem[{Liu et~al.(2024{\natexlab{d}})Liu, Li, Li, Li, Zhang, Shen, and Lee}]{llava-next}
Haotian Liu, Chunyuan Li, Yuheng Li, Bo~Li, Yuanhan Zhang, Sheng Shen, and Yong~Jae Lee. 2024{\natexlab{d}}.
\newblock \href {https://llava-vl.github.io/blog/2024-01-30-llava-next/} {Llava-next: Improved reasoning, ocr, and world knowledge}.

\bibitem[{Liu et~al.(2024{\natexlab{e}})Liu, Zheng, and Chen}]{PAI}
Shi Liu, Kecheng Zheng, and Wei Chen. 2024{\natexlab{e}}.
\newblock Paying more attention to image: A training-free method for alleviating hallucination in lvlms.
\newblock In \emph{Proc. of ECCV}, pages 125--140.

\bibitem[{OpenAI et~al.(2024)OpenAI, Hurst, Lerer, Goucher, Perelman, Ramesh, Clark, Ostrow, Welihinda, Hayes et~al.}]{GPT-4o}
OpenAI, Aaron Hurst, Adam Lerer, Adam~P. Goucher, Adam Perelman, Aditya Ramesh, Aidan Clark, AJ~Ostrow, Akila Welihinda, Alan Hayes, and 1 others. 2024.
\newblock Gpt-4o cystem card.

\bibitem[{Ren et~al.(2023)Ren, Wang, Zhu, Wang, Xiao, and Zhu}]{ren2023overcoming}
Zhibo Ren, Huizhen Wang, Muhua Zhu, Yichao Wang, Tong Xiao, and Jingbo Zhu. 2023.
\newblock Overcoming language priors with counterfactual inference for visual question answering.
\newblock In \emph{Proc. of CCL}, pages 600--610.

\bibitem[{Rohrbach et~al.(2018)Rohrbach, Hendricks, Burns, Darrell, and Saenko}]{chair}
Anna Rohrbach, Lisa~Anne Hendricks, Kaylee Burns, Trevor Darrell, and Kate Saenko. 2018.
\newblock Object hallucination in image captioning.
\newblock In \emph{Proc. of EMNLP}, pages 4035--4045.

\bibitem[{Sun et~al.(2024)Sun, Shen, Cao, Liu, Li, Shen, Gan, Gui, Wang, Yang et~al.}]{AugRLHF}
Zhiqing Sun, Sheng Shen, Shengcao Cao, Haotian Liu, Chunyuan Li, Yikang Shen, Chuang Gan, Liangyan Gui, Yu-Xiong Wang, Yiming Yang, and 1 others. 2024.
\newblock Aligning large multimodal models with factually augmented rlhf.
\newblock In \emph{Findings of ACL}, pages 13088--13110.

\bibitem[{Team et~al.(2024)Team, Georgiev, Lei, Burnell, Bai, Gulati, Tanzer, Vincent, Pan, Wang et~al.}]{Gemini-15}
Gemini Team, Petko Georgiev, Ving~Ian Lei, Ryan Burnell, Libin Bai, Anmol Gulati, Garrett Tanzer, Damien Vincent, Zhufeng Pan, Shibo Wang, and 1 others. 2024.
\newblock Gemini 1.5: Unlocking multimodal understanding across millions of tokens of context.

\bibitem[{Vaswani et~al.(2017)Vaswani, Shazeer, Parmar, Uszkoreit, Jones, Gomez, Kaiser, and Polosukhin}]{transformer}
Ashish Vaswani, Noam Shazeer, Niki Parmar, Jakob Uszkoreit, Llion Jones, Aidan~N Gomez, {\L}ukasz Kaiser, and Illia Polosukhin. 2017.
\newblock Attention is all you need.
\newblock In \emph{Proc. of NIPS}.

\bibitem[{Wang et~al.(2024{\natexlab{a}})Wang, Wu, Han, Peng, Zhong, Zhang, Dong, Li, Li, Wang et~al.}]{VIGC}
Bin Wang, Fan Wu, Xiao Han, Jiahui Peng, Huaping Zhong, Pan Zhang, Xiaoyi Dong, Weijia Li, Wei Li, Jiaqi Wang, and 1 others. 2024{\natexlab{a}}.
\newblock Vigc: Visual instruction generation and correction.
\newblock In \emph{Proc. of AAAI}, pages 5309--5317.

\bibitem[{Wang et~al.(2023{\natexlab{a}})Wang, Wang, Xu, Zhang, Gu, Jia, Wang, Xu, Yan, Zhang et~al.}]{AMBER}
Junyang Wang, Yuhang Wang, Guohai Xu, Jing Zhang, Yukai Gu, Haitao Jia, Jiaqi Wang, Haiyang Xu, Ming Yan, Ji~Zhang, and 1 others. 2023{\natexlab{a}}.
\newblock Amber: An llm-free multi-dimensional benchmark for mllms hallucination evaluation.
\newblock \emph{arXiv preprint arXiv:2311.07397}.

\bibitem[{Wang et~al.(2023{\natexlab{b}})Wang, Zhou, Xu, Shi, Zhao, Xu, Ye, Yan, Zhang, Zhu et~al.}]{wang2023evaluation}
Junyang Wang, Yiyang Zhou, Guohai Xu, Pengcheng Shi, Chenlin Zhao, Haiyang Xu, Qinghao Ye, Ming Yan, Ji~Zhang, Jihua Zhu, and 1 others. 2023{\natexlab{b}}.
\newblock Evaluation and analysis of hallucination in large vision-language models.
\newblock \emph{arXiv preprint arXiv:2308.15126}.

\bibitem[{Wang et~al.(2023{\natexlab{c}})Wang, Li, Dai, Chen, Zhou, Meng, Zhou, and Sun}]{Anchors}
Lean Wang, Lei Li, Damai Dai, Deli Chen, Hao Zhou, Fandong Meng, Jie Zhou, and Xu~Sun. 2023{\natexlab{c}}.
\newblock Label words are anchors: An information flow perspective for understanding in-context learning.
\newblock In \emph{Proc. of EMNLP}, pages 9840--9855.

\bibitem[{Wang et~al.(2024{\natexlab{b}})Wang, Bai, Tan, Wang, Fan, Bai, Chen, Liu, Wang, Ge et~al.}]{qwen2-vl}
Peng Wang, Shuai Bai, Sinan Tan, Shijie Wang, Zhihao Fan, Jinze Bai, Keqin Chen, Xuejing Liu, Jialin Wang, Wenbin Ge, and 1 others. 2024{\natexlab{b}}.
\newblock {Qwen2-vl}: Enhancing vision-language model's perception of the world at any resolution.
\newblock \emph{arXiv preprint arXiv:2409.12191}.

\bibitem[{Wang et~al.(2024{\natexlab{c}})Wang, Pan, Ding, and Biemann}]{ICD}
Xintong Wang, Jingheng Pan, Liang Ding, and Chris Biemann. 2024{\natexlab{c}}.
\newblock Mitigating hallucinations in large vision-language models with instruction contrastive decoding.
\newblock In \emph{Findings of ACL}, pages 15840--15853.

\bibitem[{Wu et~al.(2022)Wu, Zhao, Zhao, Zhang, Yuan, Zhao, and Jiang}]{wu2022overcoming}
Yike Wu, Yu~Zhao, Shiwan Zhao, Ying Zhang, Xiaojie Yuan, Guoqing Zhao, and Ning Jiang. 2022.
\newblock Overcoming language priors in visual question answering via distinguishing superficially similar instances.
\newblock In \emph{Proc. of COLING}, pages 5721--5729.

\bibitem[{Xiao et~al.(2024)Xiao, Tian, Chen, Han, and Lewis}]{StreamingLLM}
Guangxuan Xiao, Yuandong Tian, Beidi Chen, Song Han, and Mike Lewis. 2024.
\newblock Efficient streaming language models with attention sinks.
\newblock In \emph{Proc. of ICLR}.

\bibitem[{Yang et~al.(2025)Yang, Li, Cao, and Xu}]{AD-HH}
Tianyun Yang, Ziniu Li, Juan Cao, and Chang Xu. 2025.
\newblock Understanding and mitigating hallucination in large vision-language models via modular attribution and intervention.
\newblock In \emph{Proc. of ICLR}.

\bibitem[{Yang et~al.(2023)Yang, Li, Lin, Wang, Lin, Liu, and Wang}]{GPT-4v}
Zhengyuan Yang, Linjie Li, Kevin Lin, Jianfeng Wang, Chung-Ching Lin, Zicheng Liu, and Lijuan Wang. 2023.
\newblock The dawn of lmms: Preliminary explorations with gpt-4v (ision).

\bibitem[{Ye et~al.(2025)Ye, Zeng, Li, Li, and Fan}]{DetailCap}
Qinghao Ye, Xianhan Zeng, Fu~Li, Chunyuan Li, and Haoqi Fan. 2025.
\newblock Painting with words: Elevating detailed image captioning with benchmark and alignment learning.
\newblock In \emph{Proc. of ICLR}.

\bibitem[{Yin et~al.(2025)Yin, Si, and Wang}]{VAF}
Hao Yin, Guangzong Si, and Zilei Wang. 2025.
\newblock Clearsight: visual signal enhancement for object hallucination mitigation in multimodal large language models.
\newblock In \emph{Proc. of CVPR}, pages 14625--14634.

\bibitem[{Yin et~al.(2024)Yin, Fu, Zhao, Xu, Wang, Sui, Shen, Li, Sun, and Chen}]{Woodpecker}
Shukang Yin, Chaoyou Fu, Sirui Zhao, Tong Xu, Hao Wang, Dianbo Sui, Yunhang Shen, Ke~Li, Xing Sun, and Enhong Chen. 2024.
\newblock Woodpecker: Hallucination correction for multimodal large language models.
\newblock \emph{Science China Information Sciences}, 67(12):220105.

\bibitem[{Zhai et~al.(2024)Zhai, Yang, Zhao, Xu, Shen, Zhao, Keutzer, Li, Yan, and Fan}]{HallE-Switch}
Bohan Zhai, Shijia Yang, Xiangchen Zhao, Chenfeng Xu, Sheng Shen, Dongdi Zhao, Kurt Keutzer, Manling Li, Tan Yan, and Xiangjun Fan. 2024.
\newblock \href {https://openreview.net/forum?id=9Ebi1euQZQ} {Halle-switch: Rethinking and controlling object existence hallucinations in large vision-language models for detailed caption}.

\bibitem[{Zhang et~al.(2025)Zhang, Wan, Kan, Ma, Stepputtis, Ramanan, Salakhutdinov, Morency, Sycara, and Xie}]{DeGF}
Ce~Zhang, Zifu Wan, Zhehan Kan, Martin~Q Ma, Simon Stepputtis, Deva Ramanan, Russ Salakhutdinov, Louis-Philippe Morency, Katia Sycara, and Yaqi Xie. 2025.
\newblock Self-correcting decoding with generative feedback for mitigating hallucinations in large vision-language models.
\newblock In \emph{Proc. of ICLR}.

\bibitem[{Zhao et~al.(2025)Zhao, Zhang, Sun, and Feng}]{CICD}
Jianfei Zhao, Feng Zhang, Xin Sun, and Chong Feng. 2025.
\newblock Cross-image contrastive decoding: Precise, lossless suppression of language priors in large vision-language models.
\newblock \emph{arXiv preprint arXiv:2505.10634}.

\bibitem[{Zhou et~al.(2024)Zhou, Cui, Yoon, Zhang, Deng, Finn, Bansal, and Yao}]{AMOH}
Yiyang Zhou, Chenhang Cui, Jaehong Yoon, Linjun Zhang, Zhun Deng, Chelsea Finn, Mohit Bansal, and Huaxiu Yao. 2024.
\newblock Analyzing and mitigating object hallucination in large vision-language models.
\newblock In \emph{Proc. of ICLR}.

\bibitem[{Zhu et~al.(2025{\natexlab{a}})Zhu, Ji, Chen, Xu, Ye, and Liu}]{IBD}
Lanyun Zhu, Deyi Ji, Tianrun Chen, Peng Xu, Jieping Ye, and Jun Liu. 2025{\natexlab{a}}.
\newblock Ibd: Alleviating hallucinations in large vision-language models via image-biased decoding.
\newblock In \emph{Proc. of CVPR}, pages 1624--1633.

\bibitem[{Zhu et~al.(2025{\natexlab{b}})Zhu, Tao, Dong, and Xu}]{DAC}
Younan Zhu, Linwei Tao, Minjing Dong, and Chang Xu. 2025{\natexlab{b}}.
\newblock Mitigating object hallucinations in large vision-language models via attention calibration.
\newblock \emph{arXiv preprint arXiv:2502.01969}.

\end{thebibliography}
\clearpage
\appendix

\section{Two-Stage Optimization} \label{sec:alg}
We present the algorithm of our method in Algorithm~\ref{alg:attn_align}.

\section{Detailed Experiments} \label{Sec:Detailed Experiments}
\subsection{Benchmarks}
We evaluate the performance of our method across three widely adopted multimodal hallucination benchmarks on the image captioning task.
These include CHAIR, DetailCap, and AMBER.

\paragraph{CHAIR} \cite{chair} evaluates the proportion of hallucinated objects—those generated by the model but not present in the reference annotations. Following prior work, we randomly sample 500 images from the MSCOCO \cite{MSCOCO} dataset for evaluation.
CHAIRs and CHAIRi are the main metrics to evaluate hallucination:
\begin{equation}
    \begin{aligned}
    \text{CHAIRs} = \frac{|\text{Hallucinated Objects}|}{|\text{All Objects}|}, \\
    \text{CHAIRi} = \frac{|\text{Hallucinated Sentences}|}{|\text{All Sentences}|}
    \end{aligned}
\end{equation}

\paragraph{DetailCaps} \cite{DetailCap} is a fine-grained image captioning benchmark, accompanied by ground-truth detail captions generated by GPT-4V \cite{GPT-4v}, Gemini 1.5 Pro \cite{Gemini-15}, and GPT-4o \cite{GPT-4o} for evaluation purposes. It comprises 4,870 images from various datasets; we use a subset of 700 images from MSCOCO in our experiments.
CAPTURE evaluates the alignment between generated and reference captions by computing F1 scores using both hard and soft matching across three semantic aspects: entities ($F1_{\text{obj}}$), attributes ($F1_{\text{attr}}$), and relations ($F1_{\text{rel}}$). The final score is calculated as a weighted average:
\begin{equation}
\text{CAPTURE} = \frac{\alpha F1_{\text{obj}} + \beta F1_{\text{attr}} + \gamma F1_{\text{rel}}}{\alpha + \beta + \gamma}
\end{equation}
where $\alpha = 5$, $\beta = 5$, and $\gamma = 2$.

\paragraph{AMBER} \cite{AMBER} contains 1,004 carefully curated high-quality images, each with manually annotated objects.
AMBER contains multiple evaluation metrics: \emph{CHAIR}, \emph{Cover}, \emph{Hal}, and \emph{Cog}.
Given a list of annotated objects $A_{obj} = { obj_1^A, obj_2^A, \cdots, obj_n^A }$ and a set of generated objects $R'_{obj}$, each metric is defined as follows:
\begin{equation}
\begin{aligned}
\text{CHAIR} &= 1 - \frac{\mathrm{len}(R'_{obj} \cap A_{obj})}{\mathrm{len}(R'_{obj})}, \\
\text{Cover} &= \frac{\mathrm{len}(R'_{obj} \cap A_{obj})}{\mathrm{len}(A_{obj})}, \\
\text{Hal} &= \frac{|{ \text{CHAIR} > 0 }|}{|\text{All Captions}|}, \\
\text{Cog} &= \frac{\mathrm{len}(R'_{obj} \cap H_{obj})}{\mathrm{len}(R'_{obj})}
\end{aligned}
\end{equation}
where $H_{obj}$ denotes the set of hallucinated target objects generated by LVLMs, and All Captions refers to the total number of generated captions.





\subsection{Baselines}
We conduct a comparison between our method and several SOTA de-hallucination techniques:
\begin{itemize}
    \item \textbf{VCD} \cite{VCD} introduces Gaussian noise into images to activate language priors, thereby constructing negative contexts and removing these priors through contrastive decoding. However, this approach leads to a loss of visual information. Moreover, the use of noisy images deviates from the model's training distribution, potentially causing performance degradation.
    \item \textbf{ICD} \cite{ICD} constructs negative contexts by designing adversarial instructions and applies contrastive decoding to mitigate their influence. Like VCD, it faces challenges such as visual information loss and performance bias.
    \item \textbf{IBD} \cite{IBD} strengthens the model's focus on visual information by using the original context as a negative reference, and further refines contrastive decoding via inter-layer and inter-context consistency mechanisms.
    \item \textbf{VAR} \cite{VAR} reallocates attention from sink visual tokens to other visual tokens, allowing the model to capture more detailed visual information.
    \item \textbf{VAF} \cite{VAF} rebalances the attention allocation between instructions and visual inputs, redirecting attention from the textual instructions toward the visual information.
    \item \textbf{DeGF} \cite{DeGF} generates negative contexts via cross-modal back-translation and strengthens consistency signals throughout the contrastive decoding process.
    \item \textbf{CICD} \cite{CICD} leverages the consistency of language priors across images by using different images to construct negative contexts. It detects detrimental priors via consistency analysis and removes them through contrastive decoding, while retaining beneficial priors essential for accurate understanding.
\end{itemize}

\begin{algorithm*}[htbp]
\caption{Attention Distribution Alignment within a Layer}
\label{alg:attn_align}
\renewcommand{\algorithmicrequire}{\textbf{Input:}}
\renewcommand{\algorithmicensure}{\textbf{Output:}}
\renewcommand{\algorithmicreturn}{\textbf{Return:}} 
\begin{algorithmic}
\REQUIRE 
    attention weights $\mathbf{W} \in \mathbb{R}^{H \times L \times L}$, \\
    the end indexes of visual tokens $e$, \\
    hyperparameters $\kappa$ and $\omega$

\ENSURE aligned attention weights $\hat{\mathbf{W}}$

\STATE
$\mathbf{W}_{\text{cur}} \gets \mathbf{W}[:, -1, :]$  \quad \# Attention weights of the last token \\
$\mathbf{W}_{\text{sf}} \gets \operatorname{Softmax}(\mathbf{W}_{\text{cur}})$ \quad \# Apply softmax normalization \\
$\mathbf{W}_S \gets \mathbf{W}_{\text{sf}}[:, e+1:]$ \quad \# Attention to semantic representations \\
$\mathbf{W}_O \gets \mathbf{W}_{\text{sf}}[:, :e+1]$  \quad \# Attention to other representations 

\STATE
\# Head categorization: \\
$H_S \gets (\sum \mathbf{W}_S > \sum \mathbf{W}_O)$  \quad \# Semantic heads \\
$H_O \gets \neg H_S$ \quad \# Other heads \\
$H_c \gets ( \max \mathbf{W}_S > \kappa \cdot \sum \mathbf{W}_S)$ \quad \# Heads focusing on core information \\
$H_g \gets \neg H_c$ \quad \# Heads fail to focus on core information \\
$H_{S_g} \gets H_S \cap H_g$ \quad \# Global semantic heads \\
$H_{S_c} \gets H_S \cap H_c$ \quad \# Core semantic heads

\STATE
\# Two-Stage Optimization:
\IF{$|H_O| > 0$ \AND $ |H_{S_g}| > 0$}
    \STATE \# Guide the other heads using the global semantic heads \\
    $m_1 \gets \frac{1}{|H_{S_g}|} \sum \mathbf{W}_{\text{cur}}[H_{S_g}, :]$ \quad \#  Average pooling \\
    $\mathbf{W}_{\text{cur}}[H_O, :] \gets (\mathbf{W}_{\text{cur}}[H_O, :] + \omega \cdot m_1)/(1 + \omega)$
\ENDIF

\IF{$|H_{S_c}| > 0$ \AND $ |H_{S_g}| > 0$}
    \STATE \# Guide the global semantic heads using the core semantic heads \\
    $m_2 \gets \max \mathbf{W}_{\text{cur}}[H_{S_c}, :]$ \quad \# Max pooling \\
    $\mathbf{W}_{\text{cur}}[H_{S_g}, :] \gets (\mathbf{W}_{\text{cur}}[H_{S_g}, :] + \omega \cdot m_2)/(1 + \omega)$
\ENDIF

\STATE
$\mathbf{W}[:, -1, :] \gets \mathbf{W}_{\text{cur}}$ \quad \# Update attention weights

\RETURN{$\mathbf{W}$}

\end{algorithmic}
\end{algorithm*}

\begin{table}[t]
  \centering
  \begin{tabular}{lcccccc}
    \toprule
    \textbf{Setting} & \textbf{Cs}$\downarrow$ & \textbf{Ci}$\downarrow$ & \textbf{R} $\uparrow$ & \textbf{R-Cs-Ci}$\uparrow$ \\
    \midrule
    $\kappa=0.15$ & 21.6 & 7.1 & 57.6 & 28.9         \\
    \rowcolor{gray!20}
    $\kappa=0.2$ & 17.8 & 5.5 & 56.9 & \textbf{33.6}         \\
    $\kappa=0.25$ & 12.6 & 6.3 & 49.1 & 30.2        \\
    $\kappa=0.3$ & 11.0 & 5.4 & 44.8 & 28.4        \\
    \bottomrule
  \end{tabular}

  \caption{
      Explore on hyperparameter $\kappa$ with the \emph{Focused} mode.
      }\label{Tb_GL}
\end{table}

\subsection{Hyperparameters}\label{Sec_hy}
\begin{figure}
  \centering
  \includegraphics[width=0.8\linewidth]{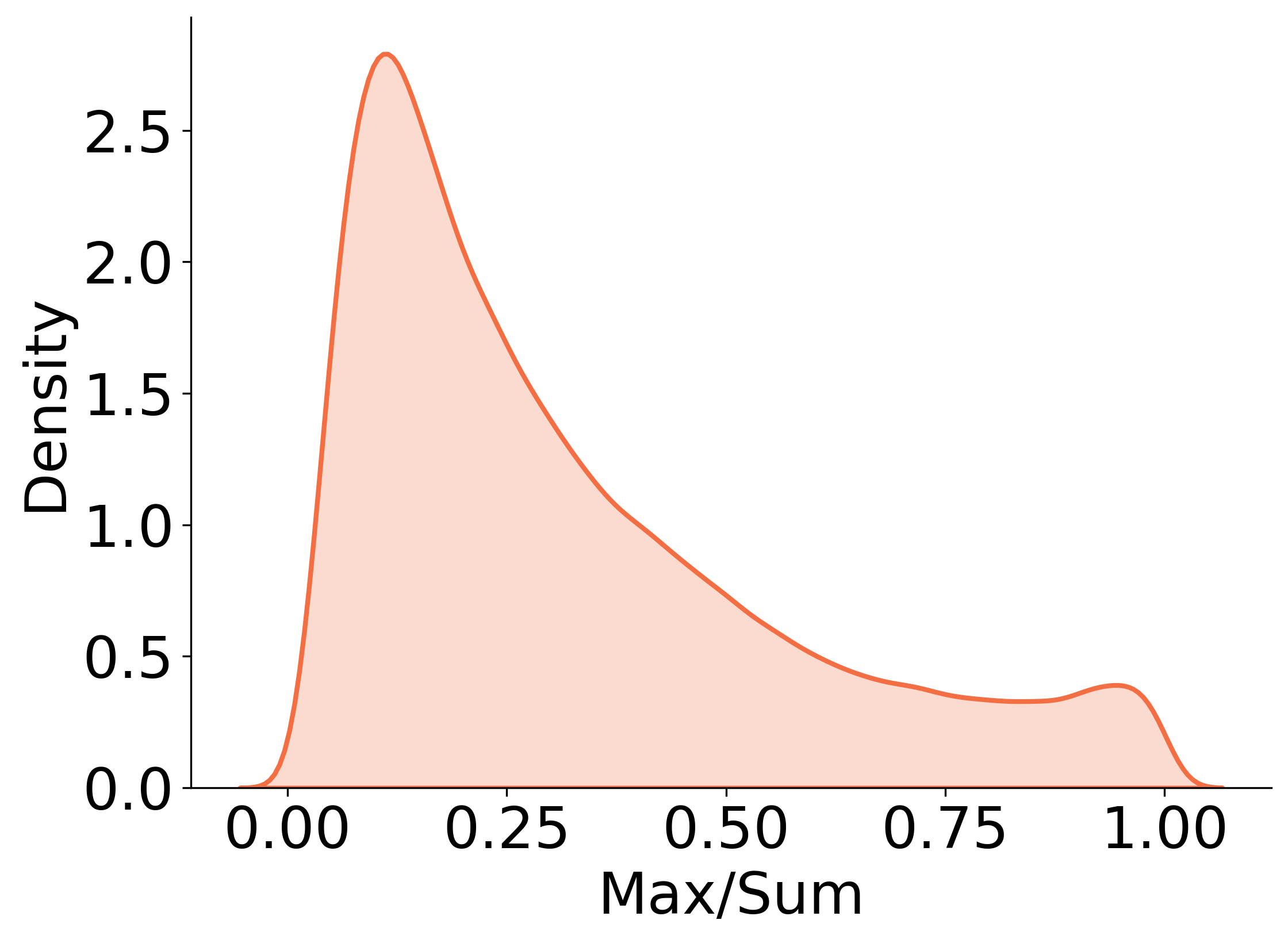}
  \caption{
  Semantic attention peaks.
    We plot the KDE (Kernel Density Estimation) of the peak attention weights for attention heads focusing on semantic representations. The x-axis represents the proportion of the highest-attended representation's attention weight among all semantic representations, while the y-axis denotes the density.
  }
  \label{Fig_gl}
\end{figure}

We investigate the appropriate setting of hyperparameters using LLaVA-1.5-7B. The parameter $\kappa$ serves as the threshold for distinguishing between core semantic heads and global semantic heads. We analyze the peak attention values of semantic heads, as shown in Fig. \ref{Fig_gl}.
Based on this analysis, we explore the impact of $\kappa$ under the focused mode, with the results presented in Tab.\ref{Tb_GL}. A higher threshold encourages the model to attend to more central information, resulting in a lower hallucination rate but also a reduced recall.
To balance these trade-offs, we define a heuristic metric to evaluate the model's overall performance. Based on the experimental results, we set $\kappa=0.2$.

The attention adjustment strength $\omega$ and the starting layer (SL) for attention optimization are interdependent. Therefore, we perform a joint search over these two hyperparameters, with the results summarized in Tab.\ref{Tb_grid}.
Based on two heuristic evaluation metrics, we select two representative hyperparameter configurations, corresponding to the Focused mode ($\omega$=4, SL=5) and Balanced modes ($\omega$=0.5, SL=9).

\begin{table*}[tbp]
  \centering
  {
  \begin{tabular}{lcccccc}
    \toprule
    \textbf{Setting} & \textbf{CHAIRs}$\downarrow$ & \textbf{CHAIRi}$\downarrow$ & \textbf{Recall} $\uparrow$ & \textbf{R-Cs-Ci}$\uparrow$ & \textbf{2R-Cs-Ci}$\uparrow$ \\
    \midrule
    $\omega$=0.5, SL=2 & 32.8 & 8.7 & 68.0 & 26.5 & 94.5         \\
    $\omega$=0.5, SL=5 &  34.4 & 9.4 & 68.8 & 25.0 & 93.8         \\
    \rowcolor{gray!20}
    $\omega$=0.5, SL=9 &  31.4 & 9.1 & 67.8 & 27.3 & \textbf{95.1}         \\
    $\omega$=0.5, SL=17 &  45.2 & 12.3 & 74.3 & 16.8 & 91.1         \\
    $\omega$=1, SL=2 & 29.4 & 7.7 & 64.4 & 27.3 & 91.7          \\
    $\omega$=1, SL=5 &  27.4 & 8.0 & 64.7 & 29.3 & 94.0         \\
    $\omega$=1, SL=9 &  26.0 & 6.7 & 63.7 & 31.0 & 94.7         \\
    $\omega$=1, SL=17 &  47.0 & 13.5 & 74.3 & 13.8 & 88.1         \\
    $\omega$=2, SL=2 &  23.4 & 7.3 & 60.5 & 29.8 & 90.3         \\
    $\omega$=2, SL=5 &   23.6 & 6.7 & 59.9 & 29.6 & 89.5        \\
    $\omega$=2, SL=9 &    23.6 & 7.1 & 61.2 & 30.5 & 91.7       \\
    $\omega$=2, SL=17 &   47.8 & 13 & 74.9 & 14.1 & 89.0        \\
    $\omega$=4, SL=2 &   18.0 & 6.1 & 55.5 & 31.4 & 86.9       \\
    \rowcolor{gray!20}
    $\omega$=4, SL=5 &   17.8 & 5.5 & 56.9 & \textbf{33.6} & 90.5        \\
    $\omega$=4, SL=9 &   21.6 & 7.1 & 57.6 & 28.9 & 86.5        \\
    $\omega$=4, SL=17 &   47.0 & 13.0 & 74.1 & 14.1 & 88.2        \\
    \bottomrule
  \end{tabular}
  }
  \caption{
    Results of grid search.
      \emph{SL} stands for \emph{Start Layer}, indicating the layer from which attention optimization begins.
      \emph{Cs}, \emph{Ci}, and \emph{R} separately represent \emph{CHAIRs}, \emph{CHAIRi}, and \emph{Recall}.
      }\label{Tb_grid}
\end{table*}

\section{Case Studies}
\begin{figure*}[tbp]
  \centering
  \includegraphics[width=\textwidth]{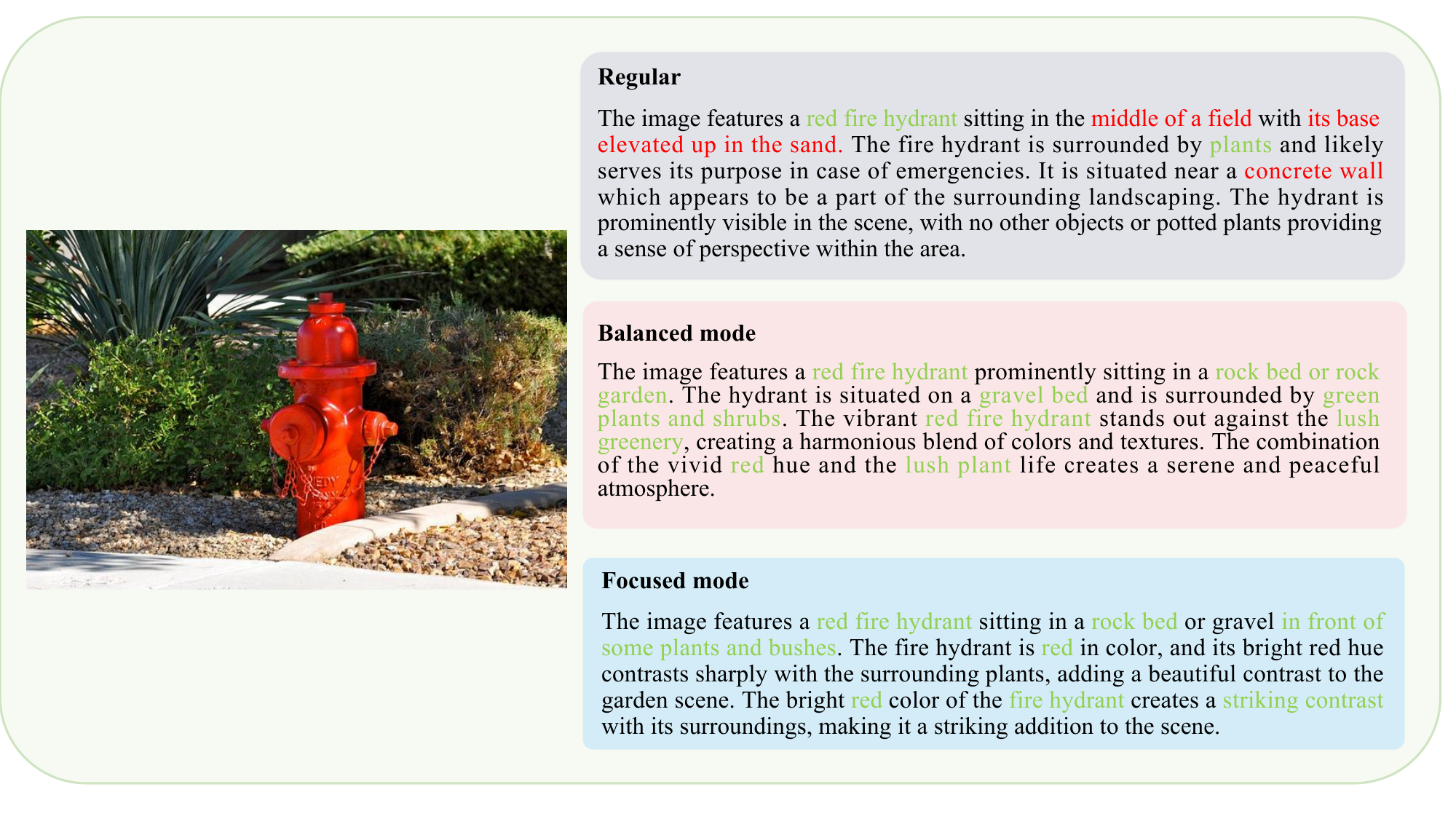}
  \caption{Case study I.}
  \label{fig:case1}
\end{figure*}

\begin{figure*}[tbp]
  \centering
  \includegraphics[width=\textwidth]{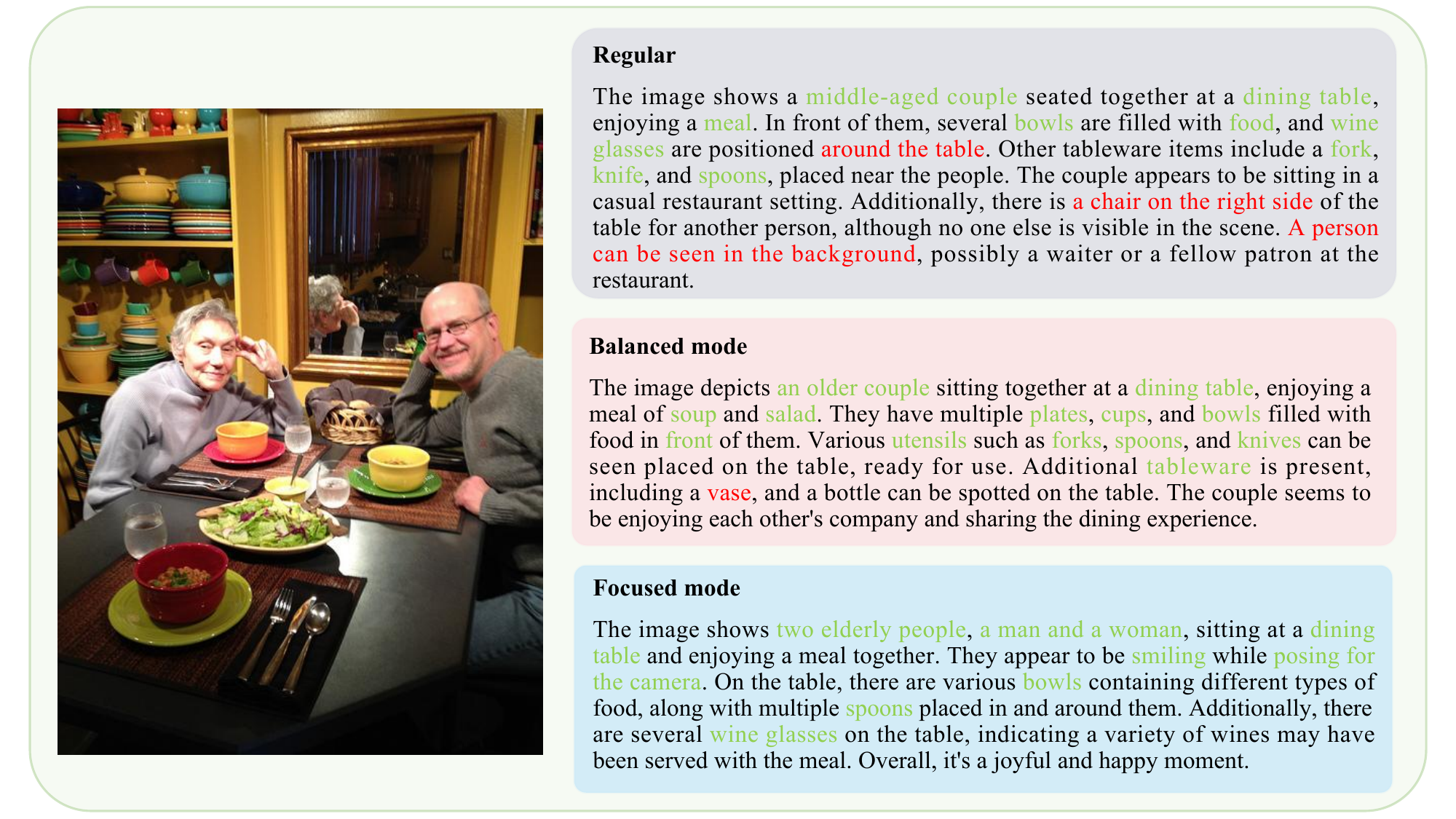}
  \caption{
  Case study II.
  }
  \label{fig:case2}
\end{figure*}

To demonstrate the effectiveness of our method in mitigating hallucinations, we provide qualitative case studies.
We select one simple image (Fig.~\ref{fig:case1}) and one complex image (Fig.~\ref{fig:case2}) as case studies. The captions were generated by LLaVA-1.5, with hallucinated content highlighted in red and image-consistent content highlighted in green.

It can be observed that Regular Decoding produces a large amount of hallucinated content, even in the simple image. In contrast, our method effectively reduces the occurrence of such hallucinations. Both the Focused and Balanced modes significantly mitigate hallucinations; however, the Focused mode makes the model overly conservative, potentially overlooking fine-grained details. For example, in Fig.~\ref{fig:case2}, the Focused mode fails to capture items such as \emph{knife} and \emph{fork}, whereas the Balanced mode not only reduces hallucinations but also preserves more detailed information.


\end{document}